\documentclass[10pt,twocolumn,letterpaper]{article}

\usepackage[pagenumbers]{cvpr} 

\usepackage{graphicx}
\usepackage{amsmath}
\usepackage{amssymb}
\usepackage{booktabs}
\usepackage[table,xcdraw]{xcolor}
\usepackage{comment}

\newcommand{\defeq}{\stackrel{\text{def}}{=}}
\newcommand{\point}[1]{\mathbf{#1}}

\newcommand{\pointing}{Pointing{`04}}
\newcommand{\OMT}{\mathrm{OMT}}
\newcommand{\dM}{d_\mathrm{M}} 
\newcommand{\Ric}{\mathrm{Ric}} 
\newcommand{\Obj}{\mathrm{Obj}}
\newcommand{\pmax}{\point{p}_{\mathrm{max}}}
\newcommand{\pmin}{\point{p}_{\mathrm{min}}}

\newlength{\twodigitswidth}
\newcommand{\fixedwidth}[1]{\makebox[\twodigitswidth][r]{#1}}

\newcommand{\markone}{*}
\newcommand{\marktwo}{$\blacktriangleright$}
\newcommand{\markthree}{$\nmid$}
\newcommand{\markfour}{ $\clubsuit$ }
\newcommand{\markfive}{$\star$ }
\newcommand{\marksix}{§}

\newtheorem{definition}{Definition}[section]

\usepackage[pagebackref,breaklinks,colorlinks]{hyperref}

\usepackage[capitalize]{cleveref}
\crefname{section}{Sec.}{Secs.}
\Crefname{section}{Section}{Sections}
\Crefname{table}{Table}{Tables}
\crefname{table}{Tab.}{Tabs.}

\begin{document}

\title{Ollivier-Ricci Curvature For Head Pose Estimation From a Single Image}

\author{Lucia Cascone\\
University of Salerno\\
Department of Computer Science\\
{\tt\small lcascone@unisa.it}
\and
Riccardo Distasi\\
University of Salerno\\
Department of Computer Science\\
{\tt\small ricdis@unisa.it }
\and
Michele Nappi\\
University of Salerno\\
Department of Computer Science\\
{\tt\small mnappi@unisa.it }
}
\maketitle

\begin{abstract}
Head pose estimation is a crucial challenge for many real-world applications, such as attention and human behavior analysis. This paper aims to estimate head pose from a single image by applying notions of network curvature.  In the real world, many complex networks have groups of nodes that are well connected to each other with significant functional roles.  Similarly, the interactions of facial landmarks can be represented as complex dynamic systems modeled by weighted graphs. The functionalities of such systems are therefore intrinsically linked to the topology and geometry of the underlying graph.  In this work, using the geometric notion of Ollivier-Ricci curvature (ORC) on weighted graphs as input to the XGBoost regression model, we show that the intrinsic geometric basis of ORC offers a natural approach to discovering underlying common structure within a pool of poses. Experiments on the BIWI, AFLW2000 and \pointing{} datasets show that the ORC\_XGB method performs well compared to state-of-the-art methods, both landmark-based and image-only.
\end{abstract}

\section{Introduction}
\label{sec:introduction}

Head pose estimation (HPE) has been a popular subject of study over the last twenty years. The pose is typically described by three angles (pitch, yaw, roll) that express face orientation with respect to a system of Cartesian axes centered on the head. Several applications benefit from understanding the face orientation in 3D space: face recognition, driver-assistance systems, virtual reality, human-robot interaction, and surveillance just to name a few \cite{chuang2014estimating} \cite{park2004semantic}. 
Critical factors such as uneven lighting, background, or occlusions (masks, scarves, glasses, etc.) can become particularly challenging issues when dealing with uncontrolled acquisitions.

It is possible to divide the different approaches used to solve this problem into two different categories: those that implement methods based on landmarks, and those that process only the raw images. Head pose estimation based on facial landmarks is quite attractive, both because  many advanced sensors now integrate landmark detection~\cite{keselman2017intel}, and because this approach is generally more robust to occlusion since it establishes a correspondence between 2D images and 3D facial models. The main limitation of this strategy is the difficulty of landmark extraction, especially for extreme poses.

\begin{figure*}[ht!]
    \centering
    \includegraphics[width=.8\textwidth]{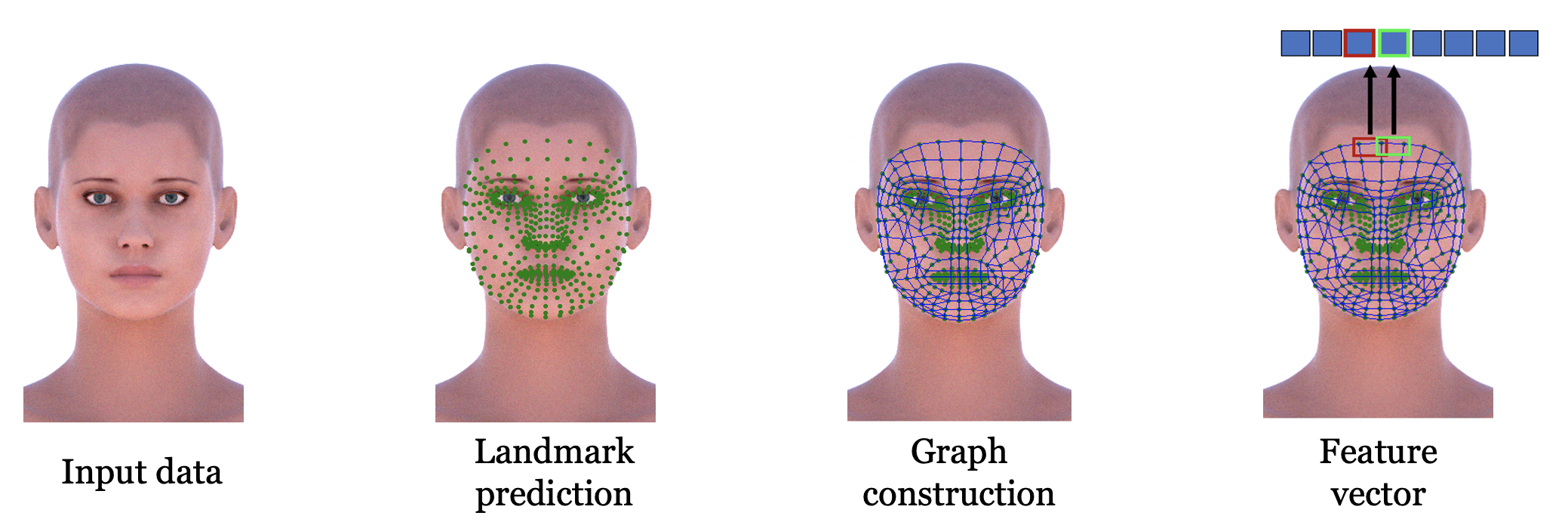}
    \caption{A graph is built using the 468 facial landmarks as vertices. The amount of Ollivier-Ricci curvature between adjacent nodes makes the feature vector.}
    \label{fig:grafo}
\end{figure*}

The idea at the foundation of this paper is to use the Ollivier-Ricci  curvature \cite{OLLIVIER2009810} (ORC) as an aid to estimate head pose by evaluating the optimal, minimum cost movement of face landmarks. We propose a connection graph that takes selected facial landmarks as vertices, on which ORC can be calculated and used as a geometric descriptor. The Ollivier-Ricci curvature distribution is shown to be different from graph to graph and can act as a graph fingerprint or graph kernel. From these premises, the question to be addressed is: Can  the  ORC curvature help us in solving  the  challenging  problem of head pose estimation? To answer this, we implemented a system whose essential parts are shown in Figure \ref{fig:grafo}.
Given an input image, the proposed system constructs a graph based on 468~facial landmarks preliminarily detected. It then regresses the head pose angle using Ollivier-Ricci curvature information for pairs of vertices. There are no other widely known methods that exploit this particular geometric descriptor for head pose estimation. The performance of the proposed method is tested on three datasets that are well known in the field. The accuracy is compared with the state of the art by replicating the same conditions and protocols so that the comparison is as meaningful as possible.  The performance obtained compares well; it actually even exceeds current accuracies, whether considering works that use landmarks or works that do not.

\section{Related work}
\subsection{Application fields of Ricci Curvature}
A notion of Ricci curvature (Ollivier-Ricci), based on optimal transportation theory, has been proposed by Yann Ollivier in \cite{OLLIVIER2009810}.
 Recently, this geometric descriptor has become a popular topic and has been applied to various fields: \textit{1)~Medicine.} In \cite{sandhu2015graph}, the authors adopt the notion of Ollivier-Ricci curvature to distinguish between cancer and normal samples.  In particular, they analyze gene co-expression networks derived from large-scale genomic studies of cancer. Cancer networks show a higher curvature than their normal counterparts. \textit{2)~Biology.} In the work \cite{doi:10.1021/acs.jcim.0c01415}, the goal is the prediction of protein-ligand binding affinity, which is a crucial aspect in drug design. Here, molecular structures and interactions are modeled as graphs on which the Ollivier-Ricci curvature is calculated. This information is then used as a descriptor for a Machine Learning (ML) model. \textit{3)~Finance.} In \cite{finance}, the authors use the Ollivier-Ricci curvature as an economic indicator for systemic risk that captures local to global system-level fragility---not only in financial markets but also in broader economic networks that include the banking ecosystem. \textit{4)~Wireless networks.} The authors in \cite{7526617} lay out a quantitative analysis that shows how the performance metrics of different routing protocols are affected by the discrete Ollivier-Ricci curvature of a network. In particular, they highlight how the concept of Ollivier-Ricci curvature can be used for wireless networks in terms of ``transport cost'', which can be related to both routing energy and queue occupancy.


\subsection{Head pose estimation}

The literature is very rich in works dealing with the HPE problem. It is possible to divide the different approaches used for solving this problem into two different categories: those that implement landmark-based methods, and those that do not use landmarks and only process simple images.

\textbf{Landmark-based methods.} In \cite{9522799}, the authors construct a landmark-connection graph and use Graph Convolutional Networks, to which they add two more models to improve performance. A different, more geometric idea on how to exploit landmarks can be found in~\cite{wsm}: the landmarks are placed in a spider's web that overlaps the face. The same idea is developed in \cite{Abate2020179}, where the feature vector is input into a regression algorithm.

\textbf{Image-based methods.} The leitmotif of these methods is to obtain a pose estimate from a simple image, without extracting any landmarks. In \cite{Bisogni20213192}, this approach is adopted without even applying ML techniques. The self-similarities present in a facial image are used to determine a fractal coding vector. This vector is compared with a reference gallery of poses to provide an estimate of pitch, yaw and roll.  A cascade ML approach is used instead in \cite{Wang2019196}. The authors define four categories of poses and use a classifier to identify which images belong to which category. A regression model refines the estimate. The same approach, i.e., to work on a single image and then apply regression, can be found also in \cite{Abate2021}, \cite{Yang20191087}, \cite{Drouard20171428},  and \cite{8945218}. In  \cite{Abate2021}, the feature vector is obtained by combining self-similar structures used for fractal image compression. By contrast, the idea in \cite{Yang20191087} is to apply a regression model to intermediate features deemed significant after learning the fine-grained structure mapping for spatial clustering of features before aggregation. Vo et al.~\cite{8945218}  use the global features extracted from a histogram of oriented gradients in a multi-stacked autoencoders neural network. In \cite{Drouard20171428}, high-dimensional feature vectors are extracted from bounding boxes of faces; the vectors are then fed into a mixture of linear regressions followed by Bayes inversion.

Deep networks have been used, too: in~\cite{Behera2021223},  an attention mechanism captures subtle changes in images. The idea in~\cite{Hu2021198} is to create fully convolutional neural networks without fully  connected layers, starting from a Bernoulli  heat map. The authors of~\cite{Gu20171531} also follow the deep learning paradigm, but rather than working on single images, dynamic facial analysis is performed by working on videos. There are several more works that use deep learning to tackle HPE, such as \cite{9435939} and~\cite{Alioua}, but a full discussion of the vast available literature is out of the scope of this paper.

\section{Methods and materials}

\begin{definition}[Graph]
A graph~$G=(V,E)$ is a finite nonempty set~$V$ of objects called \emph{vertices} or \emph{nodes}, together with a
possibly empty set~$E$ of couples of elements of~$V$ called \emph{edges} \cite{chartrand1996graphs}.
\end{definition}

\begin{definition}[Edge weighted graph]
An edge weighted graph is a pair $(G,W)$, where $G = (V, E)$ is a graph and $W : E \xrightarrow{} [0, \infty)$ is a nonnegative real function on the edges, called their \emph{weight}.
\end{definition}
An edge weighted graph is said to be \emph{undirected} if $(v_1,v_2)\in V$ implies $(v_2,v_1)\in V$, and $W(v_1,v_2)=W(v_2,v_1)$. In other words, the connection between any two nodes is symmetrical. All the graphs mentioned in this paper are undirected.

Once we have a fully connected edge weighted graph $(G, W)$, we can \emph{induce} a metric $d(\cdot,\cdot)$ on $V\times V$ by considering the minimum weight path between any two vertices:
 \begin{equation}
    \label{metrica_indotta}
    d(v_1,v_2)=
        \min
        \sum_{i=0}^n W(v_{k_i},v_{k_{i+1}}),
\end{equation}
where the minimum is obtained by considering all edge paths from $v_1$ to $v_2$---that is, over all choices
of $n$ and $k_0, \ldots k_n$ such that $v_{k_0}=v_1$, $v_{k_n}=v_2$, and $(v_{k_i}, v_{k_{i+1}})\in E$ for each~$i$.
We call $d$ the metric
\emph{induced} by the edge weight $W$.

In our context, the nodes represent actual face landmarks, coded by means of their euclidean coordinates, so the most natural choice for the weight function is a measure of their geometric distance. In particular, the choice fell on the Manhattan distance.
If $v_{k_i}\equiv(x_{k_i}, y_{k_i})$ and $v_{k_{i+1}}\equiv(y_1, y_2,...y_n)$,
their Manhattan distance~$d_M$ is defined as
\begin{equation}
    \label{metrica_manhattan}
    \dM(\point{x}, \point{y})=\sum_{i=0}^{n}|x_i-y_i|.
\end{equation}
In the Euclidean space, this is simply the squared distance---that is, the length measured along axis-aligned ``blocks''.

In this paper, the graph nodes are the facial landmarks extracted by MediaPipe \cite{MediaPipe}. MediaPipe is  a set of libraries, pre-trained models and methods for solving different types of problems. The most relevant problem in our discussion is facial landmark detection. The framework makes it possible to detect 468 facial landmarks, arranged in fixed quads and represented by their coordinates.

\subsection{Optimal Mass Transportation and Wasserstein distance: Discrete formulation}

The problem of optimal mass transportation consists in finding the most efficient transformation of a mass distribution into another. The efficiency of such a transportation is measured by a cost function~$c(\cdot, \cdot)$. In 1781 Gaspard Monge, a French mathematician, was the first to address this problem \cite{10.1112/plms/s1-14.1.139}. This was then also taken up and studied later, in 1942, by Leonid V. Kantorovich~~\cite{Kantorovich}. Kantorovich's formulation is more general and, unlike Monge's, allows masses to be divided.
Let $(X,d)$ be a complete and separable metric space (Polish space) with metric~$d$, and let $\xi$ be its Borel $\sigma$-algebra. 

Given two points~$\point{x_i}, \point{x_j} \in X$,
the cost of transporting one unit of mass
from $\point{x_i}$ to~$\point{x_j}$ is written as
$c_{i,j} \defeq c(\point{x_i}, \point{x_j})$.
This unit cost function is often
a power of the distance between the two points---in other words,
$c_{i,j} = d(\point{x_i}, \point{x_j})^r$ for some $r \geq 1$,
although different choices are also possible~\cite{schrieber2019algorithms}.
Let $\pi_{i,j}=\pi(\point{x_i}, \point{x_j})$ denote the total mass transferred from $\point{x_i}$ to~$\point{x_j}$, where $\pi$ is a transport plan finitely supported on $X \times X$ that assumes values on $[0,1]$, and let $p, q \in \xi$  be two 
probability distributions defined for $\point{x_i}$ and $\point{x_j}$’s neighbor nodes, expressed as finite Dirac sums:
\begin{equation}
    p(S_i) \defeq \sum_{{x_{ik}}\in S_i} p_k\delta_{k}(S_i) 
    \quad\text{and}\quad 
    q(S_j) \defeq \sum_{{x_{jk}}\in S_j} q_k\delta_{k}(S_j)\, 
\end{equation}
where $S_i=\{x_{i1},x_{i2},...,   x_{in}\},S_j=\{x_{j1},x_{j2},...,   x_{jm}\}$ are the finite collections of these support points,
$p_k$ and $q_k$ are constants in $[0, 1]$, and the Dirac measure $\delta_{k}$ is defined as
\begin{equation}
 \delta_{k}(S)=
 \begin{aligned}
 \begin{cases}
 1 \quad \text{if } \point{x_k} \in S\\
 0 \quad \text{if } \point{x_k} \notin S.\\
 \end{cases}
 \end{aligned}
 \end{equation}

The problem of optimal mass transport looks for an optimal transport plane~$\pi$ that minimizes the total cost of moving mass from $p$ to~$q$ \cite{villani2021topics}:
\begin{equation}
\begin{aligned}
 \OMT =
\min\sum_{i,j}c_{i,j}\pi_{i,j}
 \end{aligned}
 \label{eq:optimal_plane}
\end{equation}
subject to $\pi_{i,j} \geq 0$, $\sum_{j} \pi_{i,j}=p_i$ and
$\sum_{j} \pi_{i,j}=q_j$, for each $i=1,...,n$ and~$j=1,...,m$.
Note that both the constraints and the objective function are linear. Therefore, this is a linear programming problem.
In the common case where the cost function~$c$ is defined as a power of the metric~$d$, it is possible to define a new distance---the Wasserstein distance~$W_r(\cdot,\cdot)$---as the $r$-th root of the minimum cost value from Eq.~(\ref{eq:optimal_plane}):
\begin{equation}
    W_r(p,q)\defeq \OMT^{1/r}.
    \label{eq:wasserstein}
\end{equation}

\subsection{Ricci curvature of Riemannian manifolds}

\begin{figure}
    \centering
    \includegraphics[width=.5\textwidth]{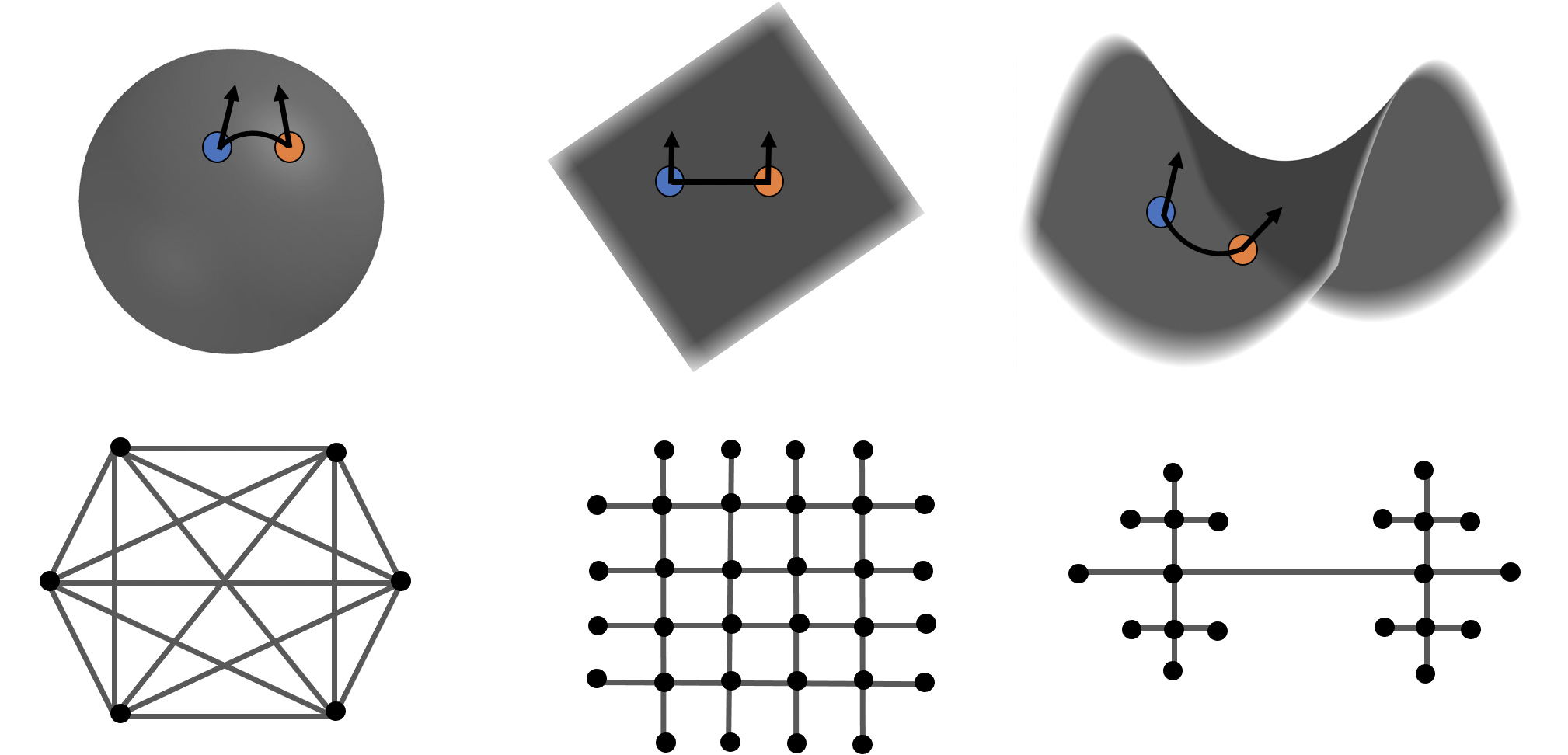}
    \caption{From left to right and from top to bottom: a sphere and a clique yields that have positive curvature, a plane and a regular grid with zero curvature and a hyperboloid with a tree-like structure that have negative curvature.}
    \label{fig:curvatura}
\end{figure}

Before we can analyze and understand the concept of curvature on graphs we need to impose some additional structure on the space \cite{hoorn2020ollivier}.  In particular, we must introduce the spaces on which graphs are defined, i.e. Riemannian manifolds. 
A manifold $M$ with a Riemannian metric is called a \emph{Riemannian manifold} and
denoted as $(M, g)$, where $g$ is the metric tensor.
\begin{definition}[Riemannian metric] Suppose for every point $\point{p}$ in a manifold $M$, an inner product $\langle\cdot,\cdot\rangle_\point{p}$ is defined on a tangent space of $M$ at~$\point{p}$, TpM. Then the
collection of all these inner products is called the Riemannian metric.
\end{definition}
Roughly speaking, the curvature of a geometric space is the local measure of how much it ``differs'' from Euclidean space. The notion of curvature, as a quantitative measure of how curved a space is, was introduced by Gauss and Riemann. In modern geometry it has become a central topic. Given a surface in 3-dimensional Euclidean space, the Gaussian curvature at a point is defined as the signed area distortion of the Gauss map sending a point on the surface to its unit normal vector.  For example, a sphere has positive curvature, a plane has zero curvature and a hyperboloid has negative one, as depicted in Fig.~\ref{fig:curvatura}.
Gauss showed that curvatures depend only on the induced Riemannian metric on the surface, i.e., they are independent of how a surface is embedded in the 3-dimensional space. The most popular curvatures are: the sectional curvature $K$ (for each point $x$ and each plane $P \subset T_xM$, the curvature $K_x(P)$ is a number), the Ricci curvature $\Ric$ (for each point $x$, the curvature $\Ric_x$ is a quadratic form on the tangent space $T_xM$), and the scalar curvature $S$ (in two dimensions, it is exactly twice the Gaussian curvature) \cite{Villani2009}. All of them are governed by the Riemannian curvature tensor $R$.

\begin{definition}[Riemannian curvature tensor]
Given an $n$-dimensional Riemannian manifold $(M,g)$ equipped with its Levi-Civita connection  $\nabla$, a point $\point{p}\in M$, and two vectors $u,v \in T_pM$, the Riemannian curvature tensor with respect to $u,v$ is a linear map  defined as
\begin{equation}
\begin{aligned}
 R(u,v):T_pM \xrightarrow{} T_pM \\
 w \xrightarrow{} R(u,v)w \\
 R\defeq \nabla_Y\nabla_X-\nabla_X\nabla_Y+\nabla_{[X,Y]}
 \end{aligned}
\end{equation}  
where in the last term we have the Lie algebra bracket.
\end{definition}
At its core, the Riemann curvature can be thought of as a tensor with four indices. It returns the measure of to what extent the manifold $M$ is non-diffeomorphic with flat Euclidean space. Of the three notions of curvature (sectional, Ricci, scalar), the sectional one is the most precise; in fact, the knowledge of all sectional curvatures is equivalent to the knowledge of the Riemann curvature.

\begin{definition}[Sectional curvature]
 Given an $n$-dimensional Riemannian manifold $(M,g)$, a point $\point{p}\in M$, and two vectors $u,v \in T_pM$, the sectional curvature is defined as
 \begin{equation}
     K(u,v)\defeq \frac{\langle R(u,v)u,v) \rangle}{\langle u,u \rangle \langle v,v \rangle - \langle u,v \rangle^2}
 \end{equation}
 where $\langle \cdot, \cdot \rangle$ denotes the inner product on the tangent space.
\end{definition}

\begin{definition}[Ricci curvature]
 Given an n-dimensional Riemannian manifold $(M,g)$, a point $\point{p}\in M$, a unit vector~$e \in T_pM$, and an orthonormal basis $(e, e_2, ..., e_n )$
 of $T_pM$, the Ricci curvature is defined as
 \begin{equation}
     \Ric_p(e,e)\defeq \frac{1}{n-1}\sum_{i=2}^{n}K(e, e_i)
 \end{equation}
 where $K(\cdot, \cdot)$ is the sectional curvature.
\end{definition}
In other words,
the Ricci curvature is obtained 
by averaging the sectional curvature over all unit vectors of $T_pM$. It should be noted that Ricci curvature controls the local behavior of geodesics: In the neighborhoods with negative curvature, the geodesics diverge; when the curvature is positive, they converge. 

Geometrically, by analyzing the Ricci curvature it is possible to control the rate of growth of the volume of a ball as a function of its radius. Furthermore, given two balls, it also allows us to investigate 
the volume of their overlap as a function of both their radii and the distance between their centers.  Obviously, there is a direct correlation between the volume of the overlap of two balls and the cost of transportation to move one to the other. A larger overlap volume corresponds to a lower transportation cost. This clears up the relationship between the Ricci curvature and the optimal transport \cite{10.1112/plms/s1-14.1.139}. An explicit formula relating these concepts was developed by Ollivier \cite{OLLIVIER2009810}.


\subsubsection{Ollivier-Ricci curvature for general metric spaces}
Wasserstein distance plays a key role in Ollivier’s approach to Ricci curvature.

\begin{definition}[Ollivier’s Ricci curvature] 
  Given an $n$-dimensional Riemannian manifold $(M,g)$ with Riemannian volume measure $\mu$, for each $x, y\in M$ the Ollivier’s Ricci curvature along the shortest path $xy$ is:
\begin{equation}
    k(x,y)=1-\frac{W(m_x,m_y)}{g(x,y)}
\end{equation}
where fixed a value $\epsilon>0$,
$m_x=\mu|_{B(x,\epsilon)}/\mu(B(x,\epsilon))$, with $B(x,\epsilon)$ the ball of radius $\epsilon$ at $x$,  and $W(m_x,m_y)$ is the Wasserstein distance with respect to the cost function $c(x,y) = g(x,y)$.
\end{definition}

This definition highlights the correlation between Ricci curvature and the optimal transport problem. Furthermore, it shows how this notion is defined for general metric spaces with probability measures.  A positive Ricci curvature corresponds to a situation where the neighbors of two centers are overlapping or otherwise close, a negative curvature implies a distance between the neighbors of two centers, and finally, zero Ollivier-Ricci curvature or even near-zero curvature indicates that locally the neighbors are embeddable in a flat surface.  Figure~\ref{fig:curvatura} illustrates this point. It can be seen that, in general, regions tend to be more densely packed at highly positive curvatures than at negative ones.

\subsubsection{Ollivier-Ricci curvature for  weighted graph}

Ollivier's definition of discrete Ricci curvature can be adapted to weighted graphs. In simple terms, given an edge, the Ollivier-Ricci curvature is computed by defining a probability distribution over each of the neighborhoods of the two nodes at its ends and then computing between the two distributions the difference between the weight of the edge and the Wasserstein distance. 
In a weighted graph $(G,W)$, we consider the associated distance function $d$ as defined in Eq.~(\ref{metrica_indotta}). To calculate the Ollivier-Ricci curvature, for each vertex we must first define the associated probability measures.
The probability distribution defined by Lin et al. \cite{10.2748/tmj/1325886283} is extended and generalized in the work of \cite{ni2019community} to further account for edge weights. Leveraging concepts from classical differential geometry they construct the probability measure as follows. Consider a network as a discretization of a smooth manifold, a Riemannian metric and a Riemannian distance should correspond to edge weight and distance. Given an edge weighted graph $(G,W)$ whose associated induced distance in the sense of Eq.~(\ref{metrica_indotta}) is $d$, the associated normalized probability measure is defined as follows:
\begin{equation}
\label{eq:prob_mes}
\begin{aligned}
m_x^{\alpha,p}(y)=
 \begin{cases}
      \alpha &\text{if } y=x\\ 
       \frac{1-\alpha}{C}\cdot \exp(-d(x, y)^r) &\text{if } y \in N(x)\\ 
       0 & \text{otherwise,}\\ 
        \end{cases}
\end{aligned}
\end{equation}
where $N(x)$
is the set of $x$'s neighbors, while
$\alpha \in [0, 1]$ indicates the share of mass that should be left on the original node.
For example, if $\alpha = 0.6$, it means that
0.6 mass units will stay in~$x$, while 0.4 units should evenly spread into $x$'s neighbors.
The power parameter~$r$ is the measure of the weight that we give to the neighbor $x_i$ of~$x$ with respect to the distance $d(x,x_i)$. The probability measure is uniform on all neighbors of $x$ when $r=0$, as suggested in \cite{10.2748/tmj/1325886283}. Instead, for large values of~$r$, the neighbors that are far away from $x$ are strongly discounted. It turns out that in most cases,  $r=1$ and $r=2$ perform best for community detection task.
The quantity $C=\sum_{x_i \in N(x)}exp(-d(x,x_i)^r)$ is a normalizing factor.

\section{Implementation details}
\label{sec:implementation_details}
This section illustrates the steps for implementing the above method.


\textbf{Input data.} Given an image as input  we use MediaPipe \cite{MediaPipe} as the landmark face detector. For each face, 466 landmarks are detected.

\textbf{Data normalization.} Given 466 landmark points $\point{p}_i =(x_i, y_i)$, we normalize all points to the same scale by:
\begin{equation}
    \widehat{\point{p}}_i=\frac{\point{p}_i-\point{p}_0}{\pmax-\pmin}
\end{equation}
where $\point{p}_0$ is a fixed vertex, while $\pmax$ and $\pmin$ are the maximum and minimum values in the graph.

\textbf{Graph construction.}
The present work has been directed at understanding if and how Ricci curvature can provide an effective descriptor for pose estimation. This is how the graphs were constructed for this experiment; the improvement of the graph structure and its optimization could be the subject of a whole new study.

Given 466 landmark locations, 
we create a graph by using the landmarks as vertices, and connecting them according to spatial proximity and empirical considerations. Once a connection strategy for the vertices is laid out, the edges are assigned individual weights according to the Manhattan distance between the relevant two vertices. The chosen configuration is the one that had the most promising initial benchmark.

At first, a preliminary study was carried out on several graphs. In these initial tests, pose estimation was obtained through the simple use of different metrics: each test image has been associated to the image coming closest to it according to one specific metric chosen from a large pool of possible distance functions, including the ones made available by the \texttt{scipy.spatial.distance} Python module~\cite{virtanen2020scipy}. This preliminary phase also included a test using Delauney triangulation~\cite{lee1980two}. However, despite the supposed rigorous sophistication of this technique, its actual performance was poor. This result can probably be ascribed to the fact that as the points (and thus the graph nodes) get denser, the unique biometric features of the individual subject overtake pose related information. The configuration chosen after these preliminary empirical tests is shown in Fig.~\ref{fig:grafo}.

\textbf{Template creation.}
Once a graph has been produced for each image, its Ollivier-Ricci curvature is computed. Every pose is associated with a vector of 446 components, representing the curvature between graph nodes. Referring to Eq.~(\ref{eq:prob_mes}), we have
$\alpha=0.5$ and $r=2$. These values resemble the process of heat diffusion.
For a proper comparison with the latest state-of-the-art work on pose estimation, we evaluated our method on three well known and challenging datasets: AFLW2000 \cite{zhu2016face}, BIWI \cite{10.1007/978-3-642-23123-0_11}, and \pointing{} \cite{gourier2004estimating}.  AFLW2000 consists of 2000 photos of faces, mostly RGB, collected from the social network Flickr. The samples in AFLW include a wide variety of poses, ethnic traits, ages, facial expressions and environmental conditions. The BIWI dataset consists of 24 videos of 20 subjects in a controlled environment. The total number of frames in the dataset exceeds 10000. The \pointing{} dataset contains faces of 15 subjects acquired in the wild, for a total of 2790 images labeled in steps of 15$^\circ$ and 30$^\circ$ in yaw and pitch between consecutive poses.  There is no roll information. Three different experimental protocols were applied depending on the dataset under consideration. 
\begin{enumerate}
\item \emph{Protocol 1}
``One left out'' strategy, using only one subject for testing, and the others as a model for carrying out the comparisons. This protocol has only been applied to BIWI, since it is the only dataset with information about the subjects. Overall, 20 experiments were carried out, rotating the test subject. In the same guise as described in other works, one more experiment was performed using a subset of only 10 subjects
\cite{Bisogni20213192} \cite{Abate2021}.

\item \emph{Protocol 2} In this case, too, only the dataset BIWI has been used. A part of the videos were used for training, and the others for testing. This protocol has been employed by several pose estimation methods with different modalities, such as RGB, depth, and time. Our method, however,
only considers a single RGB frame \cite{Yang20191087} \cite{Behera2021223}.
In particular, many researchers select 70\% of the available videos for training and the remaining 30\%
(namely, 8 videos) for testing.

\item\emph{Protocol 3} In this case, too, a split is applied on the data: training 70\%, testing 30\% \cite{Abate2020179}.
\end{enumerate}
Some works excluded extreme poses---that is, those with angles exceeding pitch $[-30^\circ, 30^\circ]$, yaw $[-45^\circ, 45^\circ ]$ and roll $[ -20^\circ, 20^\circ ]$. Therefore, for the sake of fair comparison, an additional experiment has been added to our suite, excluding from the template the poses outside these ranges. Additionally, for Protocol~1, we performed some experiments limited to a metric approach, without the use of ML.
As already sketched above while discussing graph construction, for each subject a pose was chosen according to the minimum value of one distance function picked from a pool. The best overall performer, over varying graph configurations and graph weight functions, has always been the Jensen-Shannon distance \cite{nielsen2019jensen}.

\textbf{Regression model.} After defining the protocol, a regression model has been applied to obtain an estimate of the  pose in terms of the three axes pitch, yaw and roll. We chose the Extreme Gradient Boosting Regressor (XGB), a novel  algorithm recently introduced by Chen and Guestrin \cite{chen2016xgboost}. This implementation of the gradient boosting algorithm is both efficient and effective. Its strengths include the ability to deal with overfitting, a high robustness, and a good degree of flexibility.


\section{Discussion and comparison with the state of the art }

This section illustrates an experimental evaluation of the proposed head-pose estimation strategy.
To assess our results, we report the mean absolute error (MAE)
between the estimate and the ground truth of each angle in the test set.
Our method is compared to state-of-the-art methods, both landmark-based and image-based.

For the sake of meaningful comparison, the experiments presented here only include works that use the same protocols. Furthermore, when other works limited their experiments to a subset of the poses or adopted a different ratio of training vs.\ test data, we mimicked the same choice of data and ratios.

Before commenting on our results, a clarification is in order. In recent years, particularly in the Deep Learning field, the 300W-LP dataset~\cite{zhu2016face} has been quite popular as a testbed. A few examples of papers using 300W-LP are \cite{Behera2021223},   \cite{Yang20191087}, and  \cite{Hu2021198}. A salient feature of this dataset is that, along each original image, there are several synthesized poses derived from it. Data of this kind, however, are not suitable for processing with our method, since it relies on the analysis of the relative positions of face landmarks. Synthetic deformations distort the very foundation for methods based on the correct calculation of distances and curvatures. For these reasons, we excluded 300W-LP from our experiments.

Table \ref{tab:Biwi} shows the results on the BIWI dataset using the 3~protocols. 
In all cases, the biometric component and subject peculiarities influence the performance
when considering all subjects (as opposed to one or just a few).
This can be noticed comparing Protocol~1 vs.\ Protocol~3: there is a difference of 
$\approx${}$\mbox{2-3}^\circ$ between the data in, say, rows 9 and~27.
In the case of Protocol~1, the subject whose pose is to be estimated is not in the training set; on the contrary, with Protocol~3 the same subject can be in both the training and the testing sets.
When this happens, the method seems to `recognize' the subject and associate the picture
with the picture of the same subject in the closest pose.

Comparing Protocol~1 vs Protocol~2, neither of them has images of the same subject in the training set and in the test set. The difference is that in Protocol 2 the training set contains fewer videos.  The MAE variation between the two protocols is around $1^\circ$. This shows that a larger number of subjects in the training set makes the system more robust and reliable.
As for Protocol~1, there are several papers in the literature that work on only 10 subjects out of the 20 present in the dataset, excluding the most extreme poses.

Let us consider experiment with a reduced, 10-subject dataset. The proposed ORC$\_$XGB method produces results aligned with, or slightly better than, the current alternatives. With the same number of subjects in the dataset, including the more extreme poses, our method equals or exceeds the accuracy obtained by others. This applies even to alternatives that exclude extreme poses in their tests \cite{Abate2021}. See for instance rows 1, 2, 6, and~7 in Table~\ref{tab:Biwi}.
As already stated when discussing the implementation in Section~\ref{sec:implementation_details},
this protocol has been initially studied using a simple metric in place of a ML-based regressor.
In this regard, the discrimination is effective even without recourse to Machine Learning techniques: row~7 (no ML) compares favorably to rows 1, 2, and 3~\cite{Bisogni20213192}.
The MAE and yaw figures are especially significant. When adding a ML regressor, the overall accuracy gets a boost exceeding 1$^\circ$.

\begin{figure}
    \centering
    \includegraphics[width=.5\textwidth]{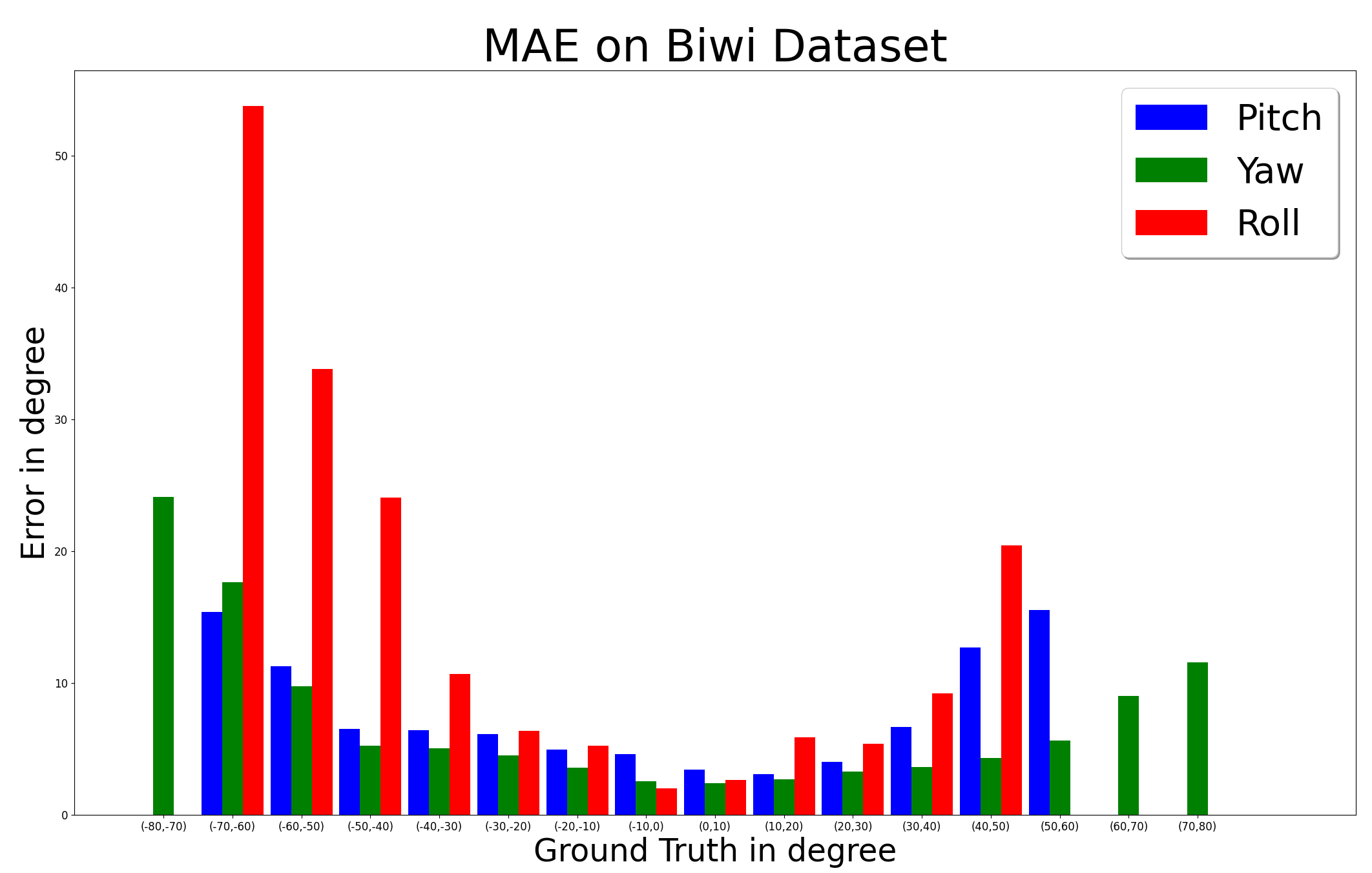}
    \caption{Errors on Biwi in terms of angular poses for pitch, yaw and roll.}
    \label{fig:MAEBiwi}
\end{figure}

When not limiting the tests to 10 subjects, but considering the whole dataset,  ORC$\_$XGB performs even better. Look at row~8 in Table~\ref{tab:Biwi}, labeled `hGLLiM'~\cite{Drouard20171428}. In this case, the proposed method improves performance in general, and particularly on the yaw axis. When ML-based regression is used, the overall performance improves by about 3$^\circ$.
\begin{table}[t] 
\begin{center}

\begin{tabular}{|l|l|l|l|l|}
\hline
\multicolumn{5}{|c|}{\textbf{Protocol 1:  one left out technique}}                                                                                                                                                                                   \\ \hline
\fixedwidth{\ } Method & Pitch                                            & Yaw                                     & Roll                                    & MAE                                     \\ \hline

\fixedwidth{1} HP$^2$IFS-LR \markthree \markone \cite{Abate2021}      & \multicolumn{1}{c|}{5.46}                                  & \multicolumn{1}{c|}{6.59}                         & \multicolumn{1}{c|}{3.8}                          & 5.28                                              \\
\fixedwidth{2}  HP$^2$IFS \markthree \markone \cite{Abate2021}         & 6.23                                                       & 4.05                                              & 3.30                                              & 4.52                                              \\
\fixedwidth{3}  FASHE \markfour \markthree \markone   \cite{Bisogni20213192}       & \multicolumn{1}{c|}{\cellcolor[HTML]{FFFFFF}4.61}          & \cellcolor[HTML]{FFFFFF}3.13                      & \cellcolor[HTML]{FFFFFF}2.74                      & 3.50                                            \\ \hline
\fixedwidth{4} \textbf{ORC\_XGB \markthree \markone}   & \multicolumn{1}{l|}{\cellcolor[HTML]{FFFFFF}\textbf{3.31}}          & \multicolumn{1}{l|}{\cellcolor[HTML]{FFFFFF}\textbf{2.30}} & \multicolumn{1}{l|}{\cellcolor[HTML]{FFFFFF}\textbf{1.76}} & \multicolumn{1}{l|}{\cellcolor[HTML]{FFFFFF}\textbf{2.45}} \\
\fixedwidth{5}  \textbf{ORC\markfour \markthree \markone}   & \multicolumn{1}{l|}{\cellcolor[HTML]{FFFFFF}\text{4.73}}          & \multicolumn{1}{l|}{\cellcolor[HTML]{FFFFFF}\text{\textbf{2.79}}} & \multicolumn{1}{l|}{\cellcolor[HTML]{FFFFFF}\text{2.82}} & \multicolumn{1}{l|}{\cellcolor[HTML]{FFFFFF}\text{3.44}} 
\\ \hline

\fixedwidth{6}  \textbf{ORC\_XGB\markone}   & \multicolumn{1}{l|}{\cellcolor[HTML]{FFFFFF}\textbf{4.06}}          & \multicolumn{1}{l|}{\cellcolor[HTML]{FFFFFF}\textbf{3.07}} & \multicolumn{1}{l|}{\cellcolor[HTML]{FFFFFF}\textbf{2.55}} & \multicolumn{1}{l|}{\cellcolor[HTML]{FFFFFF}\textbf{3.23}} \\
\fixedwidth{7}  \textbf{ORC \markfour \markone}   & \multicolumn{1}{l|}{\cellcolor[HTML]{FFFFFF}\text{5.36}}          & \multicolumn{1}{l|}{\cellcolor[HTML]{FFFFFF}\text{3.64}} & \multicolumn{1}{l|}{\cellcolor[HTML]{FFFFFF}\text{3.53}} & \multicolumn{1}{l|}{\cellcolor[HTML]{FFFFFF}\text{4.18}} 
\\ \hline

\fixedwidth{8}  hGLLiM   \cite{Drouard20171428}        & 7.65                                                       & 6.06                                              & 5.62                                              & 6.44

\\ \hline
\fixedwidth{9}  \textbf{ORC\_XGB}     & \multicolumn{1}{l|}{\cellcolor[HTML]{FFFFFF}\textbf{4.61}}          & \multicolumn{1}{l|}{\cellcolor[HTML]{FFFFFF}\textbf{3.71}} & \multicolumn{1}{l|}{\cellcolor[HTML]{FFFFFF}\textbf{3.61}} & \multicolumn{1}{l|}{\cellcolor[HTML]{FFFFFF}\textbf{3.98}} \\

\fixedwidth{10}  \textbf{ORC \markfour}   & \multicolumn{1}{c|}{\cellcolor[HTML]{FFFFFF}\textbf{6.31}}          & \multicolumn{1}{l|}{\cellcolor[HTML]{FFFFFF}\textbf{4.83}} & \multicolumn{1}{l|}{\cellcolor[HTML]{FFFFFF}\textbf{4.78}} & \multicolumn{1}{l|}{\cellcolor[HTML]{FFFFFF}\textbf{5.31}} \\

\hline

\multicolumn{5}{|c|}{\textbf{Protocol 2: split on video}}                                                                                                                                                                                       \\ \hline
\fixedwidth{\ } Method & Pitch                                            & Yaw                                     & Roll                                    & MAE                                     \\ \hline
\fixedwidth{11}  FSA-Caps-Fusion \cite{Yang20191087}   & 4.29                                                       & 2.89                                              & 3.60                                              & 3.59                                              \\

\fixedwidth{12} RGBD PRELU \marktwo \cite{7279167}   & 4.76                                                       & 5.32                                              & -                                              & 5.04                                              \\

\fixedwidth{13} RAFA-Net  \cite{Behera2021223}       & 4.30                                                       & 3.07                                              & 2.82                                              & 3.40                                              \\
\fixedwidth{14} MFDNet   \cite{9435939}           & \multicolumn{1}{l|}{\cellcolor[HTML]{FFFFFF}3.68}          & \multicolumn{1}{l|}{\cellcolor[HTML]{FFFFFF}2.99} & \multicolumn{1}{l|}{\cellcolor[HTML]{FFFFFF}2.99} & \multicolumn{1}{l|}{\cellcolor[HTML]{FFFFFF}3.22} \\
\fixedwidth{15} HR-AT-nBG  \cite{Hu2021198}      & \multicolumn{1}{c|}{\cellcolor[HTML]{FFFFFF}3.74}          & \multicolumn{1}{c|}{\cellcolor[HTML]{FFFFFF}3.07} & \multicolumn{1}{c|}{\cellcolor[HTML]{FFFFFF}3.11} & 3.31                                              \\
\fixedwidth{16} RNN $\star$ \cite{Gu20171531}        & 3.48                                              & 3.14                                              & 2.60                                              & 3.07                                              \\ 
\fixedwidth{17} EVA-GCN (vanilla)  \cite{9522799}   &    3.97                                             & 3.39                                             & 2.59                                             & 3.32                                              \\
\fixedwidth{18} {EVA-GCN}  \cite{9522799}   &    \textbf{2.82  }                                           & \textbf{2.01}                                             & {1.89}                                            & \textbf{2.24 }                                             \\

\fixedwidth{19}  VGG16  \cite{Gu20171531}    &    4.03                                           & 3.91                                           & 3.03                                           & 3.66                                              \\

\fixedwidth{20}  TriNet  \cite{cao2021vector}                                    & {3.04}           &    2.93                                         & 2.44                                          & 2.80                                              \\
\fixedwidth{21}  SSR-Net-MD  \cite{yang2018ssr}    & 4.35        &   4.24                                   & 4.19                                         & 4.26                                             \\

\hline

\fixedwidth{22} \textbf{ORC\_XGB}     & \multicolumn{1}{c|}{\cellcolor[HTML]{FFFFFF}3.51}          & \cellcolor[HTML]{FFFFFF}{2.30}             & \cellcolor[HTML]{FFFFFF}\textbf{1.86}             & \cellcolor[HTML]{FFFFFF}{2.55}             \\ \hline
\multicolumn{5}{|c|}{\textbf{Protocol 3: split on the datasets images}}                                                                                                                                                                      \\ \hline
\fixedwidth{\ } Method & Pitch                                            & Yaw                                     & Roll                                    & MAE                                     \\ \hline
\fixedwidth{23}  WSM-LgR \markthree  \cite{Abate2020179}        & \multicolumn{1}{c|}{\cellcolor[HTML]{FFFFFF}2.31}          & \cellcolor[HTML]{FFFFFF}3.12                      & \cellcolor[HTML]{FFFFFF}1.88                      & \cellcolor[HTML]{FFFFFF}2.43                      \\\hline

\fixedwidth{24}  \textbf{ORC\_XGB} \markthree       & \multicolumn{1}{c|}{\cellcolor[HTML]{FFFFFF}\textbf{1.20}}          & \cellcolor[HTML]{FFFFFF}\textbf{1.22}                      & \cellcolor[HTML]{FFFFFF}\textbf{0.95}                      & \cellcolor[HTML]{FFFFFF}\textbf{1.12 }                    \\\hline

\fixedwidth{25}  Coarse-to-Fine § \cite{Wang2019196} & 5.48                                                       & 4.76                                              & 4.29                                              & 4.84                                              \\ \hline
\fixedwidth{26} \textbf{ORC\_XGB § }     & \multicolumn{1}{l|}{\cellcolor[HTML]{FFFFFF}\textbf{1.54}}          & \multicolumn{1}{l|}{\cellcolor[HTML]{FFFFFF}\textbf{1.32}} & \multicolumn{1}{l|}{\cellcolor[HTML]{FFFFFF}\textbf{1.33}} & \multicolumn{1}{l|}{\cellcolor[HTML]{FFFFFF}\textbf{1.39}} \\
\hline
\fixedwidth{27} \textbf{ORC\_XGB}     & \multicolumn{1}{l|}{\cellcolor[HTML]{FFFFFF}\textbf{1.65}}          & \multicolumn{1}{l|}{\cellcolor[HTML]{FFFFFF}\textbf{1.45}} & \multicolumn{1}{l|}{\cellcolor[HTML]{FFFFFF}\textbf{1.38}} & \multicolumn{1}{l|}{\cellcolor[HTML]{FFFFFF}\textbf{1.49}} \\
\hline
\end{tabular}
\end{center}
\caption{Performance comparison with state-of-the-art results on the BIWI dataset for the 3~protocols. 
`\markone': only 10 subjects;  `\marksix':  training\slash{}testing ratio 80\slash{}20; `\markfive': temporal information; `\marktwo': depth information; `\markthree': without extreme poses;  `\markfour': without ML.}
\label{tab:Biwi}
\end{table}
Figure~\ref{fig:MAEBiwi} illustrates the error on the BIWI, Protocol~1, in terms of angular poses for pitch, yaw and roll. For some ranges, especially the extreme ones, the error grows noticeably. Some investigation shows that the main reason resides with the detector: when the error is larger, the landmarks themselves are off. This suggests that, if the landmark detection process were performing correctly, the overall performance could be significantly better. Even with the undermining factor of a faulty detection process, the final results are mostly acceptable. This consideration reinforces the conclusion that Ollivier-Ricci curvature works well as a descriptor for pose estimation.

The results in Table~\ref{tab:Biwi} show that the method behaves well with Protocol~2, too. The MAE drops well below 3$^\circ$.
The results on the pitch axis are somewhat worse than those shown in~\cite{Gu20171531}, although this could be at least partially due to their working with videos: the temporal information that can be extrapolated
probably helps to track head movements, improving the accuracy. Focusing on rows 17 and~18, the work in~\cite{9522799} performs better than ORC\_XGB, particularly on the pitch axis. Moreover, when the base method (EVA\_CGN vanilla) is integrated by an adaptive channel attention module and a densely connected architecture, results improve on the yaw axis as well.

Protocol~3 is perhaps the least interesting in this context, since there is significant interference introduced by individual traits vs.\ pose differences. The performance of ORC$\_$XGB still ranks at the top, with the best error figures. When excluding poses with extreme roll values, the MAE value is less than~1$^\circ$.
When using the whole dataset, there is an improvement exceeding 3$^\circ$. The results are particularly good on the pitch axis. Even when the training set is not enlarged (e.g., 80\% in~\cite{Wang2019196}), the system holds its own when compared with current methods. It is interesting to note that a smaller training set seems to have a lesser effect on accuracy for the roll axis.

Table~\ref{tab:Pointing} reports the results for the \pointing{} dataset under Protocol~1 and Protocol~3. 
When running Protocol~1, the proposed system compares well with \cite{Alioua}, \cite{Drouard20171428}, and \cite{PATACCHIOLA2017132}. The overall MAE is worse than in~\cite{8945218}, but ORC$\_$XGB produces the only value of yaw error below 7$^\circ$. This experiment shows the effectiveness of adding a regression module. As expected, excluding extreme poses improves the results, even more so than with other methods. Focusing on the comparison with \cite{Bisogni20213192} and~\cite{wsm}, which do not use ML, shows that a descriptor based on Ricci curvature can be quite effective even when simply used as a metric, without a regression module.
The data about Protocol~3 show good results in different conditions, such as a 50\slash{}50 training\slash{}testing ratio, exclusion of extreme poses. The results with this dataset are generally better for the pitch axis.

\begin{table}[t!] 
\begin{center}
\begin{tabular}{|l|l|l|l|}
\hline
\multicolumn{4}{|c|}{\textbf{Protocol 1:  one left out technique}}                                                                                                                                                 \\ \hline
Method             & Pitch                                                      & Yaw                                                        & MAE                                                        \\ \hline
\fixedwidth{1} hGLLiM \cite{Drouard20171428}           & 8.47                                                       & 7.93                                                       & 8.20                                                      \\
\fixedwidth{2} Patacchiolla (CNs) \cite{PATACCHIOLA2017132} & 10.71                                                      & 7.74                                                       & 9.22                                                       \\ 
\fixedwidth{3} SP+LPF  \cite{Alioua}           & 10.86                                                      & 7.52                                                       & 9.19                                                       \\ 
\fixedwidth{4} SAE-XGB  \cite{8945218}          & \textbf{6.16}                                                       & 7.17                                                       & \textbf{6.66}                                                       \\ \hline
\fixedwidth{5} \textbf{ORC\_XGB}       & \multicolumn{1}{l|}{\cellcolor[HTML]{FFFFFF}\text{7.16}} & \multicolumn{1}{l|}{\cellcolor[HTML]{FFFFFF}\textbf{6.64}} & \multicolumn{1}{l|}{\cellcolor[HTML]{FFFFFF}\text{6.90}} \\

\fixedwidth{6} \textbf{ORC \markfour}       & \multicolumn{1}{l|}{\cellcolor[HTML]{FFFFFF}\text{9.74}} & \multicolumn{1}{l|}{\cellcolor[HTML]{FFFFFF}\text{8.84}} & \multicolumn{1}{l|}{\cellcolor[HTML]{FFFFFF}\text{9.29}} \\

\hline
\fixedwidth{7} FASHE \markfour \markthree  \cite{Bisogni20213192}                & \multicolumn{1}{c|}{9}                                     & \multicolumn{1}{c|}{6.60}                                   & 7.80                                                        \\ 

\fixedwidth{8} WSM \markfour \markthree  \cite{wsm}                & \multicolumn{1}{c|}{6.34}                                  & \multicolumn{1}{c|}{10.63}                                 & 8.48                                                       \\ 

\fixedwidth{9} SVR \markthree \cite{Bisogni20213192}                & \multicolumn{1}{c|}{12.82}                                     & \multicolumn{1}{c|}{11.25}                                   & 	12.03                                                      \\ \hline

\fixedwidth{10} \textbf{ORC\_XGB \markthree }       & \multicolumn{1}{l|}{\cellcolor[HTML]{FFFFFF}\textbf{5.18}} & \multicolumn{1}{l|}{\cellcolor[HTML]{FFFFFF}\textbf{3.51}} & \multicolumn{1}{l|}{\cellcolor[HTML]{FFFFFF}\textbf{4.34}} \\ 
\fixedwidth{11} \textbf{ORC \markfour \markthree }       & \multicolumn{1}{l|}{\cellcolor[HTML]{FFFFFF}\textbf{8.55}} & \multicolumn{1}{l|}{\cellcolor[HTML]{FFFFFF}\textbf{4.75}} & \multicolumn{1}{l|}{\cellcolor[HTML]{FFFFFF}\textbf{6.65}} \\ 

\hline

\multicolumn{4}{|c|}{\textbf{Protocol 3:  split on the datasets images}}  \\ \hline
Method             & Pitch                                                      & Yaw                                                        & MAE                                                        \\ \hline
\fixedwidth{12} Patacchiolla (CNs) § \cite{PATACCHIOLA2017132} & 8.06                                                      & 6.93                                                       & 7.49                                                       \\ \hline
\fixedwidth{13} \textbf{ORC\_XGB §}       & \multicolumn{1}{l|}{\cellcolor[HTML]{FFFFFF}\textbf{6.73}} & \multicolumn{1}{l|}{\cellcolor[HTML]{FFFFFF}\textbf{6.89}} & \multicolumn{1}{l|}{\cellcolor[HTML]{FFFFFF}\textbf{6.81}} \\ \hline

\fixedwidth{14} WSM-LgR \markthree  \cite{Abate2020179}            & 7.55                                                       & 4.44                                                       & 5.99                                                       \\ \hline

\fixedwidth{15} \textbf{ORC\_XGB \markthree }       & \multicolumn{1}{l|}{\cellcolor[HTML]{FFFFFF}\textbf{4.79}} & \multicolumn{1}{l|}{\cellcolor[HTML]{FFFFFF}\textbf{3.76}} & \multicolumn{1}{l|}{\cellcolor[HTML]{FFFFFF}\textbf{4.27}} \\ 
\hline
\fixedwidth{16} \textbf{ORC\_XGB}       & \multicolumn{1}{l|}{\cellcolor[HTML]{FFFFFF}\textbf{6.33}} & \multicolumn{1}{l|}{\cellcolor[HTML]{FFFFFF}\textbf{6.65}} & \multicolumn{1}{l|}{\cellcolor[HTML]{FFFFFF}\textbf{6.49}} \\ 
\hline
\end{tabular}
\end{center}
    \caption{Performance comparison with state-of-the-art results on \pointing{} dataset, with respect than the Protocol~1 and Protocol~3.
`\marksix':  training\slash{}testing ratio 50\slash{}50; `\markthree': without extreme poses;  `\markfour': without ML.}
    \label{tab:Pointing}
\end{table}

With the AFLW2000 dataset, two sets of experiments were performed with Protocol~3. The first set replicates the conditions described in \cite{Abate2020179},  \cite{Abate2021}, and \cite{wsm} by excluding the more extreme poses, while the second set alters the training\slash{}testing ratio. With this dataset, too, the overall error is under 3$^\circ$. In this case, it is the roll axis that shows the best results, reaching below~2$^\circ$.
Even including the extreme poses and reducing the training percentage to~50\%, our results on pitch and particularly yaw surpass other methods, such as~\cite{Abate2021}.

\begin{table}[t!] 
\begin{center}
\begin{tabular}{|l|c|l|l|l|}
\hline
\multicolumn{5}{|c|}{\textbf{Protocol 3: split on the datasets images}}                                                                                                                                                                        \\ \hline
Method       & \multicolumn{1}{l|}{Pitch}            & Yaw                                                        & Roll                                                       & MAE                                                        \\ \hline
\fixedwidth{1} WSM \markfour \markthree  \cite{wsm}          & 4.82                                  & \multicolumn{1}{c|}{3.11}                                  & \multicolumn{1}{c|}{2.25}                                  & 3.39                                                       \\

\fixedwidth{2} WSM-BRR \markthree  \cite{Abate2020179}          & 4.67                                 & \multicolumn{1}{c|}{3.82}                                  & \multicolumn{1}{c|}{2.49}                                  & 3.66 \\

\fixedwidth{3} HP$^2$IFS-LsR \markthree \cite{Abate2021} & 6.90                                   & \multicolumn{1}{c|}{6.70}                                   & \multicolumn{1}{c|}{4.48}                                  & 6.02                                                       

                 \\ \hline
\fixedwidth{4} \textbf{ORC\_XGB \markthree} & \cellcolor[HTML]{FFFFFF}\textbf{3.51} & \multicolumn{1}{l|}{\cellcolor[HTML]{FFFFFF}\textbf{3.02}} & \multicolumn{1}{l|}{\cellcolor[HTML]{FFFFFF}\textbf{1.94}} & \multicolumn{1}{l|}{\cellcolor[HTML]{FFFFFF}\textbf{2.82}} 
\\ \hline

\fixedwidth{5} HP$^2$IFS § \markthree   \cite{Abate2021}  & \multicolumn{1}{l|}{7.46}             & 6.28                                                       & 5.53                                                       & 6.42                                                       \\ \hline
\fixedwidth{6} \textbf{ORC\_XGB} § \markthree& \cellcolor[HTML]{FFFFFF}\textbf{3.51} & \multicolumn{1}{l|}{\cellcolor[HTML]{FFFFFF}\textbf{2.93}} & \multicolumn{1}{l|}{\cellcolor[HTML]{FFFFFF}\textbf{1.96}} & \multicolumn{1}{l|}{\cellcolor[HTML]{FFFFFF}\textbf{2.80}} 
\\  \hline
\fixedwidth{7} \textbf{ORC\_XGB} & \cellcolor[HTML]{FFFFFF}\textbf{6.01} & \multicolumn{1}{l|}{\cellcolor[HTML]{FFFFFF}\textbf{4.24}} & \multicolumn{1}{l|}{\cellcolor[HTML]{FFFFFF}\textbf{5.81}} & \multicolumn{1}{l|}{\cellcolor[HTML]{FFFFFF}\textbf{5.35}} 
\\

\hline
\end{tabular}
\end{center}
\caption{Performance comparison with state-of-the-art results on AFLW2000 dataset, Protocol~3. `\markthree': without extreme poses;  `\marksix':  training\slash{}testing ratio 80\slash{}20;  `\markfour': without ML.}
\label{tab:AFLW2000}
\end{table}

\section{Conclusion and Future Work}
This paper addressed the problem of the head pose estimation with a Ollivier-Ricci curvature based method. We present a method that exploits 468 facial reference points to construct a graph on which the geometric curvature based descriptor is applied. For each pair of edges, a curvature value is computed and inserted into a feature vector. The vector is in turn fed to a regression model---the Extreme Gradient Boosting Regressor. Experiments were performed on three well known datasets, with images collected in both controlled and wild environments. The experimental results show this approach performs well when compared to state-of-the-art methods under identical protocols conditions. Therefore, Ollivier-Ricci curvature is empirically proved to be a good descriptor for this type of problem. As a future development, we plan on investigating the graph creation phase more systematically. In particular, we would like to find a geometry to improve the spatial relationship between the landmarks.  A better designed graph is likely to further improve the discriminating performance.

{\small
\bibliographystyle{ieee_fullname}
\bibliography{egbib}
}

\end{document}